\pdfoutput=1

\documentclass[10pt,twocolumn,letterpaper]{article}

\usepackage{authblk}
\usepackage{wacv}
\usepackage{times}
\usepackage{epsfig}
\usepackage{graphicx}
\usepackage{amsmath}
\usepackage{amssymb}
\usepackage{mathtools, cuted}
\usepackage{lipsum, color}
\usepackage{multirow}
\usepackage{bbm}
\usepackage{algorithm}
\usepackage{color,soul}
\usepackage[noend]{algpseudocode}
\usepackage[inline]{enumitem}
\usepackage[font=small,labelfont=bf,tableposition=top]{caption}
\usepackage{float}

\usepackage{hyperref}

\wacvfinalcopy 

\ifwacvfinal\fi
\begin{document}

\title{In Defense of the Triplet Loss Again: Learning Robust Person Re-Identification with Fast Approximated Triplet Loss and Label Distillation}


\newcommand*\samethanks[1][\value{footnote}]{\footnotemark[#1]}
\author[1]{\vspace{-2em} Ye Yuan}
\author[1]{Wuyang Chen}
\author[2]{Yang Yang}
\author[1]{Zhangyang Wang}
\affil[ ]{\vspace{-2em}} 

{\makeatletter
\renewcommand\AB@affilsepx{, \protect\Affilfont}
\makeatother
\affil[ ]{\textit {\{ye.yuan, wuyang.chen, atlaswang\}@tamu.edu}} 
\affil[ ]{\textit {yang.yang2@walmart.com}}
}

{\makeatletter
\renewcommand\AB@affilsepx{, \protect\Affilfont}
\makeatother
\affil[1]{Texas A\&M University}
\affil[2]{Walmart Technology} 
}

\affil[ ]{\small \url{https://github.com/TAMU-VITA/FAT}}
\renewcommand\Authands{ and }

\maketitle
\ifwacvfinal\thispagestyle{empty}\fi

\begin{abstract}   
The comparative losses (typically, triplet loss) are appealing choices for learning person re-identification (ReID) features. However, the triplet loss is computationally much more expensive than the (practically more popular) classification loss, limiting their wider usage in massive datasets. Moreover, the abundance of label noise and outliers in ReID datasets may also put the margin-based loss in jeopardy. This work addresses the above two shortcomings of triplet loss, extending its effectiveness to large-scale ReID datasets with potentially noisy labels. We propose a fast-approximated triplet (\textbf{FAT}) loss, which provably converts the point-wise triplet loss into its upper bound form, consisting of a point-to-set loss term plus cluster compactness regularization. It preserves the effectiveness of triplet loss, while leading to linear complexity to the training set size. A label distillation strategy is further designed to learn refined soft-labels in place of the potentially noisy labels, from only an identified subset of confident examples, through teacher-student networks. We conduct extensive experiments on three most popular ReID benchmarks (Market-1501, DukeMTMC-reID, and MSMT17), and demonstrate that FAT loss with distilled labels lead to ReID features with remarkable accuracy, efficiency, robustness, and direct transferability to unseen datasets.
\end{abstract}

\vspace{-1em}
\section{Introduction}

\begin{figure}[ht]
\begin{center}
    \includegraphics[width=.8\linewidth]{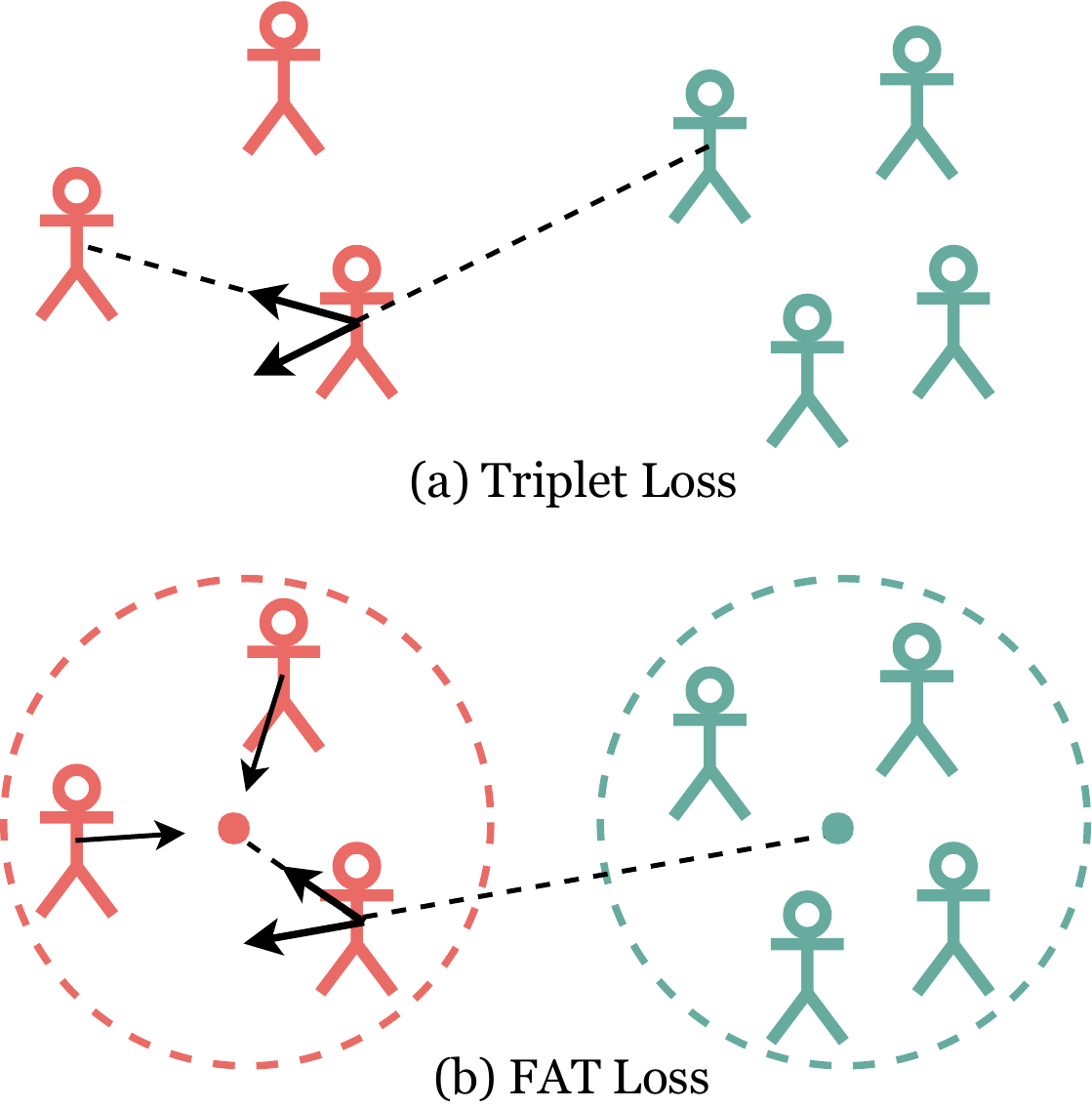}
\end{center}
\vspace{-1em}
\caption{Illustrative comparison of standard triplet loss and FAT loss. The former compares point-to-point distances, while the latter compares point-to-set distances while regularizing all cluster sets to be compact. The solid arrows depict the ``push and pull'' effect of triplet loss and the point-to-set term of FAT loss. The dash arrows represents the compactness regularization of FAT loss. See details in Section 3.}
\vspace{-1em}
\label{fig:fat}
\end{figure}

Person re-identification (ReID) has attracted tremendous attention owing to its vast applications in video surveillance, public safety, and so on. Given a person image spotted by one camera, ReID aims to accurately match that probe image against a large amount of gallery images, taken by other cameras and timestamps. The dramatic visual appearance variations of the same person, as caused by different poses, view angles, illuminations, and backgrounds, constitute serious challenges for learning robust identity representations. 

Most existing ReID algorithms use a classification loss to train their feature learning backbones \cite{zheng2016person, xiao2016learning, xiao2016end, li2017person, chen2019abd, yuan2019calibrated}. However, ReID is essentially an  ``open-ended'' retrieval problem rather than closed-set classification, \textit{e.g.}, the training and testing sets usually have no overlapped identity classes. The learned feature extractor should be able to generalize to matching unseen identities. The testing performance is evaluated by the precision and recall of the matching instances, rather than classification accuracy. Therefore, the classification-driven learning could be misaligned with the end goal. Instead, the comparative losses \cite{Schroff_2015_CVPR, deng2017marginal, ming2017simple, zheng2018centralized}, which compares the distances between two sample pairs, are naturally better choices, as empirically validated by a handful of works \cite{liu2017end, li2017person, Chen_2017_CVPR, xiao2017margin, cheng2016person}. Among many, the triplet loss \cite{Alexander_2017}, which maximizes the margin between the intra-class distance and the inter-class distance, has been mostly used in ReID, in order to explicitly embed the relative orders between right and wrong matches (\ie, the correct matches should always be closer to the query than the wrong ones).

However, an important downside of triplet loss lies in its \textbf{computational expensiveness}, which prohibits its wide usage in the large-scale ReID applications. A naive triplet loss that compares every possible pair of training samples will incur cubic complexity w.r.t. the training set size \cite{Alexander_2017}. Also, triplet loss relatively quickly learns to correctly map most trivial triplets, rendering a large fraction of all triplets uninformative. Applying triplet loss with randomly selected triplets can accelerate training but quickly stagnates, or becomes difficult to converge. Hard sample mining \cite{Yu_2018_ECCV, zhao2018adversarial} has recently become the standard practice in using triplet loss, to select only ``informative'' (a.k.a. hard) pairs rather than all pairs to enforce the loss. However, it runs the risk of causing sample bias \cite{Yu_2018_ECCV}, and often appears fragile to outliers. The vanilla triplet loss needs to calculate over all $PK(K-1)(PK-K)$ possible triplets, where $K$ denotes average number of images per identity and $P$ identities in total \cite{Alexander_2017}. The time complexity can be reduced to $PK(PK-1) + PK$ when hard sample mining is used.

In this paper, we will propose a new fast-approximated triplet (\textbf{FAT}) loss to trim down the computational cost of triplet loss without hampering its effectiveness. Viewing all images belonging to the same identity class as a cluster, the proposed FAT loss re-defines a triplet to include an anchor, its corresponding cluster centroid, and the centroid of another cluster. 
The main idea of FAT loss is to replace point-to-point distances with point-to-cluster distances, through an upper bound relaxation of the triplet form. Such a relaxation simultaneously requires the query to be closest to its ground-truth-cluster centroid, and enforces each cluster to have a compact radius. The FAT loss thus has a linear complexity w.r.t. the training set size. 

Another downside of triplet loss, as well as many other margin-based losses, lies in their \textbf{fragility to label noise}.   
Unfortunately, ReID datasets are notorious to have many noisy labels and outliers, such as label flipping, mislabel, and multi-person coexistence, due to the tedious manual annotation process. The proposed FAT loss can alleviate the label noise to some extent, by averaging all samples within the same cluster. To provide further improved robustness, we consider a distillation network to first generate soft pseudo labels for each sample, associated with its confidence. Then we use those soft labels in place of the original labels to feed into the FAT loss, where each individual sample’s contribution to the model update will be re-weighted by their label confidence. 

In sum, we strive to make triplet loss a more effective, efficient, and robust choice for ReID, via multi-fold efforts:
\begin{itemize}[noitemsep,topsep=2pt]
    \item We propose a fast-approximated triplet (FAT) loss to remarkably improve the efficiency over the standard triplet loss, with linear complexity to the training set size. It is derived by relaxing triplet loss to its upper bound form, and operates without hard sample mining.
     \item We are the first to demonstrate that explicitly considering and handling label noise can further boost ReID performance. A  distillation network is presented to assign soft labels for samples in place of the original (potentially noisy) hard labels. Combined with FAT loss, a more robust re-ID feature can be learned. 
    \item We conduct extensive experiments on three most popular ReID benchmarks, and demonstrate that FAT loss with learned soft labels lead to comparable or superior ReID performance than using triplet loss and other state-of-the-art baselines, with remarkably higher efficiency than triplet loss. We also observe improved robustness and direct transferability to unseen data. 
\end{itemize}

\section{Related Work}
\noindent{\bf Triplet Loss and Hard Sample Mining}
The triplet loss was first introduced in FaceNet \cite{Schroff_2015_CVPR}  by Google to train face embeddings for the recognition task, where softmax cross entropy loss failed to handle a variable number of classes. The goal of triplet loss is to maximize the inter-class variation while minimizing the intra-class variation. Triple loss is formulated as \eqref{triplet_loss} below, where the triplet is defined as an anchor sample $a$, a positive sample $p$ from the same class and a negative sample $n$ from a different class ($y_a$, $y_p$, $y_n$ denote class labels for $a,p,n$, respectively): 
\begin{equation} 
L_{\text{tri}} =\sum_{\substack{a,p,n \\ y_p=y_a \\ y_n \not = y_a}} 
\max \{d(a, p) + m - d(a, n), 0\}
\label{triplet_loss}
\end{equation} 
FaceNet picked a random negative for every pair of anchor and positive, which was very time-consuming. Later on, \cite{Alexander_2017} improved the efficiency of triplet loss for the ReID task, by proposing two triplet selection strategies: batch all and batch hard. The batch all strategy selects all valid triplets and averaged the loss. The batch hard strategy selects the hardest positive and negative samples within the batch when
forming the triplets. The author suggested that batch hard strategy with soft margin to yield better performance. \cite{Yu_2018_ECCV} found that selecting the hardest triplets often led to bad local minima. They argued that the bias in the triplet selection degraded the performance of learning with triplet loss, and proposed a new variant of triplet loss that adaptively corrects the distribution shift on the selected triplets. 

Besides, there are many other successful practices in applying triplet loss to ReID task. \cite{Cheng_2016_CVPR} proposed a multi-channel convolutional neural network to learn global-local parts features and improved the triplet loss requiring the intra-class feature distances to be less than a predefined threshold. \cite{Chen_2017_CVPR} extended the triplet loss to a quadruplet form and required the intra-class variations to be smaller than any inter-class variations. \cite{Yu_2018_ECCV_2} generalized the point-to-point (P2P) triplet loss to the point-to-set (P2S) form by assuming a positive set (to which the anchor belongs) and a negative set (including all other clusters) for each anchor. It then penalizes the difference between the distance from the anchor to the positive set centroid and the anchor-to-negative-centroid distance. The model was also trained in a soft hard-mining scheme with greater weights to harder samples.

Being related to previous works \cite{do2019theoretically,Alexander_2017,Yu_2018_ECCV_2}, FAT loss differs substantially in the following ways:

\begin{itemize}[noitemsep,topsep=2pt]
    \item FAT loss has linear time complexity w.r.t training dataset size: $\mathcal{O}(PK)$ or $\mathcal{O}(PK^2)$ (depending on the choice of negative set), where $K$ denotes the average image number per identity and $P$ the number of identities. Previous triplet losses have either cubic (vanilla) and quadratic (with hard sample mining) time complexity w.r.t training dataset size.
    \item FAT loss is analytically derived from the upper bound of standard triplet loss. It consists of a P2S loss term and intra-class compactness regularization. Up to our best knowledge, all previous approximations or accelerations for triplet loss, e.g., \cite{Cheng_2016_CVPR,Yu_2018_ECCV_2}, are only empirical.
    \item We studied different choices of the negative cluster/centroid, and compared their impacts. 
    Note that 
    FAT loss chooses the negative on ``cluster'' level, and does not refer to any individual sample mining. 
\end{itemize}

\noindent{\bf Learning from Noisy Labels} 
The growing scale of training datasets embrace the potential of a more powerful model, but introduces sample outliers and label noise during data collection and annotation. \cite{Wang_2018_ECCV} observed that a face recognition model trained with only a subset 30\% manually cleaned-label samples can achieve comparable performance with models trained on the full dataset. To overcome the negative effect of noisy labels, \cite{Reed_2014} proposed a bootstrap technique to modify the labels on-the-fly by augmenting the prediction objective with a notion of consistency. \cite{Liu_2016_TPAMI} extended  \cite{Natarajan_2013_NIPS} and proposed a re-weighting method that can be combined with any surrogate loss function for classification, to handle class-conditional random label flipping. \cite{Sukhbaatar_2014} introduced an extra noise layer to absorb the label noise by adapting the network outputs to the noisy label distribution. \cite{Goldberger_2016} further augmented the correction architecture by adding a softmax layer on top to explicitly connect the correct labels to noisy ones. \cite{Patrini_2017_CVPR} provided a forward-and-backward loss correction method given a class-condition label flipping probability. \cite{Vahdat_2017_NIPS} proposed a generic conditional random field (CRF) model as a robust loss to be plugged into any existing network for label space smoothness and therefore noise resistance. \cite{Wang_2018_CVPR} designed a Siamese network to distinguish clean labels from noisy labels and to simultaneously give clean labels more emphasis.

Interestingly, various label noise, such as class-conditional or sample-conditional label flipping, mislabeling, and multi-person co-existence, are extensively found in ReID dataset. Yet to our best knowledge, few previous works have formally studied how to handle them, and how that may improve ReID performance. 

\begin{algorithm*}[!ht]
\caption{Derivation of FAT loss as an upper bound for triplet loss (\ref{triplet_loss}).}\label{fat_derive}
\vspace{-1em}
\begin{equation*}
\begin{aligned}
L_{\text{tri}} &= \max\{0, d(a, p) + m - d(a, n)\} \\
& \leq \max\{0, d(a, c_a) + d(c_a, p) + m - \max\{0, d(a, c_n) - d(c_n, n)\} \} \triangleright \text{refer to both inequalities in (\ref{triangle}) \qquad \qquad \qquad} \\ 
&= \max\{0, d(a, c_a) + d(c_a, p) + m - d(a, c_n) + \min\{d(c_n, a), d(c_n, n) \} \} \triangleright \text{move d(a, $c_n$) out of inner $\max$ then reverse sign} \\
&= \max\{0, d(a, c_a) + m - d(a, c_n) + d(c_a, p) + \min\{d(c_n, a), d(c_n, n)\}\} \\
&= \max\{0, d(a, c_a) + m - d(a, c_n)\} + d(c_a, p) + \min\{d(c_n, a), d(c_n, n)\} \triangleright \text{move non-negative sums out of $\max$ \qquad \qquad} \\
& \leq \max\{0, d(a, c_a) + m - d(a, c_n)\} + d(c_p, p) + d(c_n, n) \triangleright \text{$c_a$ = $c_p$; $\min$\{d($c_n$, a), d($c_n$, n)\} $\le$ d($c_n$, n)} \\
& \leq 
\underbrace{\max\{0, d(a, c_a) + m - d(a, c_n)\}}_{\text{\textbf{anchor-dependent point-to-set loss}}} + \underbrace{R(a) + R(n)}_{\text{\textbf{cluster compactness}}}  \triangleright   \text{R() defines the radius of the cluster adnd can be pre-computed}
\end{aligned}
\vspace{-1em}
\end{equation*}
\end{algorithm*}

\noindent{\bf Network Distillation}
Network distillation was first developed in \cite{Hinton_2015} to transfer the knowledge in an ensemble of models to a single model,  using a soft target distribution produced by the former models. \cite{Bulo_2016_ICML} used distillation to train a more efficient and accurate predictor. \cite{lopez_2015} unified distillation and privileged information into one generalized distillation framework to learn better representations. \cite{Radosavovic_2018_CVPR} further extended data distillation to omni-supervised learning by ensemble of predictions from multiple transformations of unlabeled data to generate new training annotations using a single network. \cite{Luo_2018_ECCV,Garcia_2018_ECCV} applied data distillation to multi-modal training, while the testing sets might have noisy or missing modalities. As a relevant work, \cite{Li_2017_ICCV} argued that noisy labels contains useful "side information" and shall not be discarded. The authors proposed a distillation approach to learn from noisy data guided by a knowledge graph. 

Our proposed distillation algorithm to learn from noisy labels differs from previous ones in the following respects:
\begin{itemize}[noitemsep,topsep=2pt]
    \item We are free from the assumption of the existence of a manually-cleaned set. Instead, we train the teacher network with the entire noisy dataset, but only use most confident samples within a batch to update the parameters. We observed that the model updated based on a subset of confident samples can achieve similar or better performance, compared to the model trained with all noisy-labeled samples.

    \item We investigate different loss functions for distillation; the teacher network is trained with cross entropy loss with the purpose of providing pseudo soft label associated with a confidence; the student network is trained with FAT loss using the soft pseudo labels generated by the teacher network. Hence instead of mimicking a similar softmax classifier as the teacher network, the student network has the capability to ``innovate'' on a different task with the help of FAT loss, and eventually outperforms the teacher network.
\end{itemize} 

\vspace{-0.5em}
\section{Method}
% ...

\subsection{Fast Approximated Triplet (FAT) Loss}
Given an anchor image $a$ with the identity label $y_a$, the triplet loss attempts to find a positive sample $p$ with the same identity label $y_p = y_a$ and a negative sample $n$ with a different label $y_n \not = y_a$, and then maximizes the difference of distances between the positive pair $d(a,p)$ and the negative pair $d(a,n)$ by a margin $m$. We typically use the euclidean distance (or cosine similarity) between learned ReID features $f_E(a), f_E(p), f_E(n)$ as distance metrics.

However, computing triplet loss exhaustively over all possible pairs is too expensive to be practical. We propose a relaxation of the triplet loss \ref{triplet_loss} into its upper bound form. We first have the following two triangle inequalities:
\begin{equation} 
\begin{aligned}
& \max\{0, d(a, c_a) - d(c_a, p)\} \leq d(a, p) \leq d(a, c_a) + d(c_a, p) \\
& \max\{0, d(a, c_n) - d(c_n, n)\} \leq d(a, n) \leq d(a, c_n) + d(c_n, n) \\
\end{aligned} 
\label{triangle}
\end{equation}
where $c_a$, $c_n$ are defined as the centroids (average) of the clusters that $a$, $n$ belong to, respectively. Their proofs are self-evident, given that $d()$ is a well-defined distance function in some metric space. Notice that although we use Euclidean distance for $d()$ by default, our derivations are applicable to other distances too. 

We next expand our derivation as in the outline (\ref{fat_derive}). Interestingly, the upper bound consists of two terms: a point-to-set (P2S) term which depends on the anchor point; plus a penalty term on the cluster compactness, defined as the largest cluster ``radius'' among all clusters, whose value is decided by the entire dataset and is agnostic to the anchor. We minimize this upper bound instead, and name it as the \textit{fast approximated triplet} (\textbf{FAT}) loss:
\begin{equation}
\begin{aligned}
\vspace{-1em}
L_{\text{FAT}} = \sum_{\substack{a, n \\ n \not = y_a}} \max \{0, d(a, c_a) + m - d(a, c_n)\} + R(a) + R(n).
\end{aligned}
\label{fat}
\end{equation} 
As the name suggests, the new loss will give rise to similarly competitive ReID performance compared to the full triplet loss, but with tremendously better efficiency. We now analyze FAT loss w.r.t. the triplet loss from two aspects. 

As can be obviously seen from its form, FAT loss greatly accelerates the cubic/quadratic time complexity of computing triplet loss, to linear complexity, w.r.t. the training set size. We also examine how tight it approximates the original tripelet loss. Observing (\ref{fat_derive}), three relaxations take place in the second, sixth and seven lines. For the first one, the equality in (\ref{triangle}) could be met when: $a, c_a, p$ are co-linear with $a, p$ on the same side of $c_a$; while $a, c_n, n$ are also co-linear with $a, n$ on different sides of $c_n$. The second relaxation becomes tight if and only if $d(a, c_n) \ge d(n, c_n)$, which implies that $a$ is sufficiently far away from the cluster of $c_n$. For the last one, the exact equality can only be taken in a very special case, when every cluster has the same radius and every sample in a cluster distributes on a circle. In sum, when clusters are well-separated and balanced in size, FAT loss can provide a relatively tighter approximation for triplet loss. However, it is always reasonable to expect that minimizing this upper bound would lead to suppressing the original triplet loss value too.

\vspace{-1em}
\paragraph{Normalized FAT Loss}
As a margin loss, FAT loss, as well as triplet loss, is sensitive to input scales. Given the fact that ReID features are also scale-sensitive: neighboring features in the normalized space can be far away from each other in the original feature space, the learned feature are often normalized before feeding into the evaluation metrics. That could be reflected in a normalized FAT loss:
\begin{equation}
\begin{aligned}
L_{\text{FATnorm}} = & \max\{0, d(\frac{a}{||a||}, c'_a) + m - d(\frac{a}{||a||}, c'_n)\} \\
& + R'(a) + R'(n),
\end{aligned}
\label{fat-n}
\end{equation} 
where $R'$ is similarly defined as the radius of the normalized sample set. In practice, we empirically find that adding a cross entropy (CE) loss $L_\text{CE}$ term will help stabilize training with FAT or Normalized FAT loss notably. That leads to minimizing a hybrid loss ($L_{\text{CE-FAT}}$ can be replaced to $L_{\text{FAT-N}}$; $\lambda$ is a scalar):
\begin{equation} 
L_{\text{CE-FAT}} = L_{\text{FAT}}  + \lambda*L_{\text{CE}}
\label{fat-h}
\end{equation}

\begin{figure}[t]
\begin{center}
    \includegraphics[width=\linewidth]{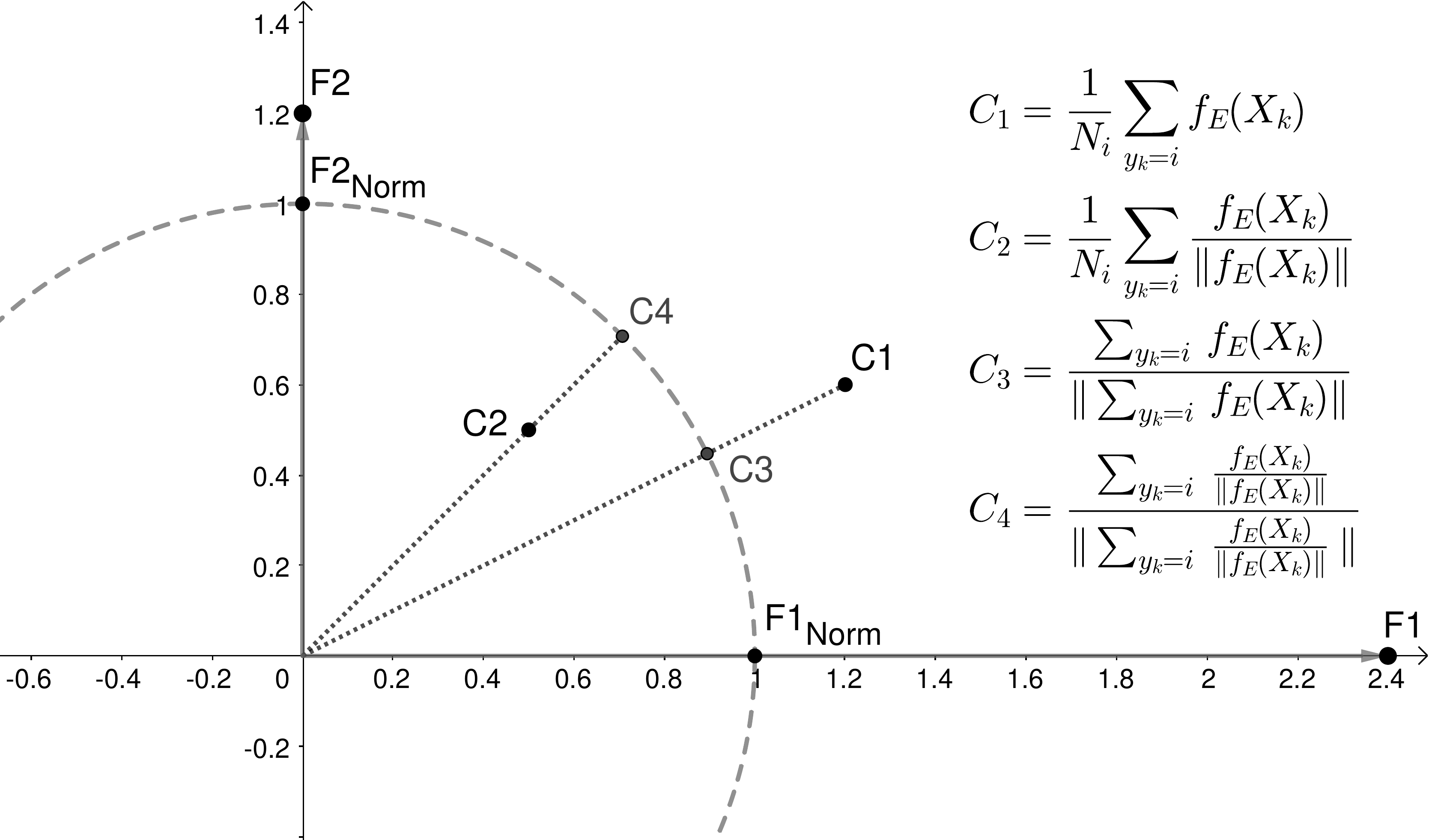}
\end{center}
\vspace{-0.5em}
\caption{Example of four different centroid options.}
\vspace{-1em}
\label{fig:centroid}
\end{figure}

\paragraph{Choices of Centroids}
The choice of cluster centroids is also found to be critical to the effectiveness of FAT loss. Four options of cluster centroids are available: i) mean of cluster features; ii) mean of normalized cluster features; iii) normalized mean of cluster features; and iv) normalized mean of normalized cluster features. Mathematically:
\begin{equation}
\begin{aligned}
& C_{i1} = \frac{1}{N_i} \sum_{y_k=i} f_E(X_k), \qquad C_{i2} = \frac{1}{N_i} \sum_{y_k=i} \frac{f_E(X_k)}{\|f_E(X_k)\|} \\
& C_{i3} = \frac{\sum_{y_k=i} f_E(X_k)}{\|\sum_{y_k=i} f_E(X_k)\|}, \quad  C_{i4} = \frac{\sum_{y_k=i} \frac{f_E(X_k)}{\|f_E(X_k)\|}}{\|\sum_{y_k=i} \frac{f_E(X_k)}{\|f_E(X_k)\|}\|} \\
\end{aligned}
\label{center}
\end{equation} 
A visual comparison of the four options are in Figure \ref{fig:centroid}.

Since the original FAT loss (\ref{fat}) is calculated based on un-normalized features, only the first centroid option $C_{i1}$ makes sense for it. The remaining three options can all be utilized for the normalized FAT loss (\ref{fat-n}). Our experiments indicate that the normalized mean of normalized cluster features $C_{i4}$ works best with the normalized FAT loss.

\subsection{Distillation for Noisy Label Robustness}
Typically, there are three common label noises in ReID datasets: i) label flip, i.e., an image is assigned to a wrong identity class; ii) mislabeling, i.e., an image does not belong to any known identity class; iii) multiple identities co-exist in one image. Similar to other margin-based losses, triplet loss is highly sensitive to label noise. Since the proposed FAT loss has a P2S term where all samples within the same cluster are averaged, hence alleviating noisy labels to some extents. We hereby propose a label distillation approach based on a teacher-student model, to improve FAT loss robustness to label noise further, using ``soft labels'' predicted from another teacher model, trained with a loss that is less sensitive to label noise, \textit{e.g.}, cross entropy. The pipeline is 
plotted in the supplementary, with details explained below.

We first use a self-bootstrapping approach to learn the teacher model robustly. The teacher net is first trained with cross entropy loss on classifying all samples (including noisy labels) for 5 epochs. It was previously observed that the network would be more inclined to learning with high confidence for ``easy samples'', within the early stage of training \cite{li2017self,katharopoulos2018not}. Those confident, easy samples are hypothesized to have labels that are semantically consistent and correct, less confusing and ambiguous, and therefore more reliable. We identify those most confidently predicted samples based on the entropy of their currently predicted softmax vectors. We then resume training for another 5 epochs; but now in each epoch, we will keep using those identified confident samples, while not using or only partially using the others that are more likely to contain label noise or outliers. We periodically repeat the above process, and each time we may gradually enlarge the pool of confident examples as the training continues. More details will be presented in Section 4.1.

After the teacher model is trained, its predictions are treated as soft labels to replace the original labels, for training the student model with FAT loss. Only the ``confident'' labels eventually selected by the teacher net will participate in averaging to estimate the cluster centroids. If we use the hybrid FAT loss (\ref{fat-h}), then soft labels are the prediction targets for the cross entropy (softmax) loss too.

\section{Experiment}

\subsection{Datasets and Implementation}

We evaluate the proposed method on three most popular large-scale benchmarks: Market-1501 \cite{Zheng_2015_ICCV}, DukeMTMC-reID \cite{Ristani_2016_ECCV, Zheng_2017_ICCV}, and MSMT17 \cite{Wei_2018_CVPR}.

{\bf Market-1501} comprises 32,668 labeled images of 1,501 identities captured by six cameras. Following \cite{Zheng_2015_ICCV}, 12,936 images of 751 identities are used for training, while the rest are used for testing. Among the testing data, the test probe set is fiCEd with 3,368 images of 750 identities. The test gallery set also includes 2,793 additional distractors.

{\bf DukeMTMC-reID} is a subset of the DukeMTMC dataset \cite{Ristani_2016_ECCV} for person ReID. This dataset contains 36,411 images of 1,812 identities, cropped from the videos every 120 frames. These images are captured by eight cameras, among which, 1,404 identities appear in more than two cameras and 408 identities (distractors) who appear in only one camera. The 1,404 identities are randomly divided, with 702 identities for training and the others for testing. In the testing set, one query image for each ID in each camera is chosen for the probe set, while all remaining images including distractors are in the gallery. 

{\bf MSMT17} is the current largest publicly-available ReID dataset. It has 126,441 images of 4,101 identities captured by 15-camera network (12 outdoor, 3 indoor). We follow the training-testing split of \cite{Wei_2018_CVPR}.The video is collected with different weather conditions at three time slots (morning, noon, afternoon). All annotations, including camera IDs, weathers and time slots, are available. MSMT17 is \textbf{significantly more challenging} than the other two, due to its massive scale, more complex and dynamic scenes, and severe \textbf{label noise} (see examples in the supplementary).

\begin{table*}[!ht]
\caption{Evaluation results on Market1501 and transfer results from Market1501 to DukeMTMC-reId. We use Resnet50 as our default backbone and trained on Market1501, with only one exception indicated by \textbf{*} using DenseNet161 backbone. }
\begin{center}
\resizebox{0.95\textwidth}{!}{
\begin{tabular}{l|c|c|c|c|c|c|c|c|c|c}
\hline \multicolumn{3}{c|}{Settings} & \multicolumn{4}{c|}{Test on Market1501} & \multicolumn{4}{c}{Transfer to DukeMTMC-reID} \\
\hline loss & negative & margin & top1 & top5 & top10 & mAP & top1 & top5 & top10 & mAP  \\
\hline\hline
Histogram Loss \cite{ustinova2016learning} & NA & NA & 59.5 & 80.7 & 86.9 & - & - & - & -  \\
Multi-loss class \cite{li2017person} & NA & NA & 83.9 & - & - & 64.4 & - & - & - & -\\
Point to Set Similarity \cite{zhou2017point} & NA & NA & 70.7 & - & - & 44.3 & - & - & - & -\\
Triplet loss \cite{Alexander_2017} & NA & 1 & 84.9 & 94.2 & - & 69.1 & - & - & - & -\\
Support Neighbor Loss \cite{li2018support} & NA & NA & 88.3 & - & - & 73.4 & - & - & - & -\\
CycleGAN \cite{Deng_2018_CVPR} & NA & NA & - & - & - & - & 38.5 & 54.6 & 60.8 & 19.9 \\
\hline
CE-FAT&ctrdAll&1&89.1&95.0&96.7&71.6&34.4&51.5&57.6&18.9\\
CE-FAT&ctrdAvg&1&89.2&95.3&97.0&72.4&35.1&51.2&57.6&19.2\\
CE-FAT&ctrdHM&1&87.1&94.7&96.3&69.9&34.3&50.8&56.9&18.0\\
CE-FAT&batchNeg&1&89.4&95.6&97.1&73.1&37.3&52.3&58.4&20.3\\
\hline
CE-P2S&ctrdAll&1&87.4&95.0&96.7&68.9&27.6&42.9&50.0&14.1\\
CE-P2S&batchNeg&1&87.2&94.6&96.7&67.0&28.1&42.6&49.2&14.3\\
\hline
CE-P2Snorm&batchNeg&0.1&87.5&95.3&96.8&68.1&27.8&41.7&48.7&13.6\\
CE-FATnorm&batchNeg&0.1&88.6&95.1&96.7&69.7&35.0&50.6&57.4&18.9\\
\hline
\hline
\textbf{CE-FAT*} (DenseNet161) & batchNeg  &1 & \textbf{91.4} & \textbf{96.6} & \textbf{97.7} & \textbf{76.4} & \textbf{40.8} & \textbf{57.1} & \textbf{63.2} & \textbf{23.4} \\
\hline
\end{tabular}}
\end{center}
\vspace{-1em}
\label{table:hyperparameters_market}
\end{table*}

\begin{table*}[!ht]
\caption{Evaluation results on DukeMTMC-reID and transfer results from DukeMTMC-reID to Market1501. We use Resnet50 as our backbone, and trained on DukeMTMC-reID, with only one exception indicated by \textbf{*} using DenseNet161 backbone.}
\begin{center}
\resizebox{0.9\textwidth}{!}{
\begin{tabular}{l|c|c|c|c|c|c|c|c|c|c}
\hline \multicolumn{3}{c|}{Settings} & \multicolumn{4}{c|}{Test on DukeMTMC-reID} & \multicolumn{4}{c}{Transfer to Market1501} \\
\hline loss & negative & margin & top1 & top5 & top10 & mAP & top1 & top5 & top10 & mAP  \\
\hline\hline
Deep-Person \cite{bai2017deep} & NA & NA & \textbf{80.9} & - & - & \textbf{64.8} & - & - & - & - \\
CycleGAN \cite{Deng_2018_CVPR} & NA & NA & - & - & - & - & 48.1 & 66.2 & 72.7 & 20.7 \\
\hline
CE-P2Snorm&batchNeg&0.1&76.5&87.3&90.6&57.3&46.5&63.9&71.0&19.9\\
CE-FATnorm&batchNeg&0.1&77.9&87.8&91.4&58.3&49.8&65.8&73.2&21.2\\ 
CE-P2S&batchNeg&1&78.2&88.5&91.8&59.5&47.0&64.6&71.4&19.7\\
CE-FAT&batchNeg&1&78.8&88.7&91.5&60.8&49.1&67.1&73.9&21.8\\
\hline
\hline
\textbf{CE-FAT}* (DenseNet161) &batchNeg&1 & 80.8 & \textbf{89.5} & \textbf{92.0} & 63.1 & \textbf{54.7} & \textbf{70.8} & \textbf{77.4} & \textbf{25.2} \\
\hline
\end{tabular}}
\end{center}
\vspace{-1em}
\label{table:hyperparameters_duke}
\end{table*}

\begin{table*}[!ht]
\caption{Evaluation results on MSMT17, DukeMTMC-reID, and Market1501. We use ResNet50 as our backbone and trained on MSMT17 with different negative sets.}
\begin{center}
\resizebox{0.96\textwidth}{!}{
\begin{tabular}{c|c|c|c|c|c|c|c|c|c|c|c|c|c}
\hline
loss & negative set & \multicolumn{4}{c|}{Test on MSMT17} & \multicolumn{4}{c|}{Transfer to DukeMTMC-reID} & \multicolumn{4}{c}{Transfer to Market1501} \\ 
\hline
PDC \cite{Su_2017_ICCV} & NA & 58.0 & 73.6 & 79.4 & 29.7 & - & - & - & - & - & - & - & - \\
GLAD \cite{Wei_2017_ACM} & NA & 61.4 & 76.8 & 81.6 & 34.0 & - & - & - & - & - & - & - & - \\
HHL \cite{Zhong_2018_ECCV} & NA & - & - & - & - & 45.0 & 59.4 & 64.4 & 23.0 & \textbf{56.0} & \textbf{75.8} & \textbf{81.2} & 26.7\\
\hline
CE-P2Snorm&batchNeg&64.8&78.3&83.0&33.8&49.1&64.9&70.6&29.2&51.6&68.9&75.5&23.9\\
CE-FATnorm&batchNeg&66.2&79.4&83.7&33.1&\textbf{51.2}&\textbf{66.1}&71.1&29.5&54.8&70.9&76.5&25.1\\
CE-P2S&batchNeg&65.2&78.5&82.9&33.7&49.9&67.6&\textbf{74.5}&22.9&48.7&63.5&69.3&\textbf{28.5}\\
\hline
CE-FAT&ctrdAll&68.8&81.4&85.4&39.1&50.9&65.0&70.2&\textbf{30.7}&51.5&69.4&75.9&24.4\\
CE-FAT&ctrdAvg&67.0&80.2&84.6&37.4&45.0&61.7&67.0&25.4&48.3&65.6&73.0&21.5\\
CE-FAT&ctrdHM&67.7&80.2&84.5&36.2&50.1&64.4&70.2&28.4&48.4&66.0&72.5&21.5\\
CE-FAT&batchNeg&\textbf{69.4}&\textbf{81.5}&\textbf{85.6}&\textbf{39.2}&49.2&64.8&69.6&28.7&50.6&68.0&74.9&23.6\\
\hline
\end{tabular}}
\end{center}
\vspace{-1.5em}
\label{table:hyperparameters_msmt}
\end{table*}

\begin{table*}
\caption{Evaluation results of the Teacher Net on MSMT17, DukeMTMC-reID, and Market1501. We use ResNet50 as our backbone and trained on MSMT17.}
\begin{center}
\resizebox{0.93\textwidth}{!}{
\begin{tabular}{c|c|c|c|c|c|c|c|c|c|c|c|c}
\hline
\multirow{2}{*}{Method} & \multicolumn{4}{c|}{Test on MSMT17} & \multicolumn{4}{c|}{Tranfer to DukeMTMC-reID} & \multicolumn{4}{c}{Tranfer to Market1501} \\ \cline{2-13} 
 & top1 & top5 & top10 & mAP & top1 & top5 & top10 & mAP & top1 & top5 & top10 & mAP \\ \hline\hline
whole set & \textbf{65.1} & 78.2 & 82.8 & \textbf{34.5} & 48.2 & 63.8 & 69.9 & 29.0 & 51.1 & 68.3 & 74.2 & 23.5 \\
\hline
hard threshold & 64.5 & 77.8 & 82.2 & 33.7 & 46.5 & 62.8 & 69.0 & 27.4 & 49.9 & 66.2 & 73.3 & 23.0 \\ 
soft threshold & 64.8 & \textbf{78.3} & \textbf{83.0} & 34.2 & 48.2 & 63.5 & 69.0 & 28.9 & 49.6 & 67.3 & 74.1 & 23.1 \\
\hline
hard percentage & 64.2 & 77.5 & 82.1 & 34.2 & 49.3 & 64.4 & 69.8 & 29.8 & 52.0 & 69.2 & \textbf{76.5} & \textbf{24.8} \\ 
soft percentage & 62.9 & 76.1 & 80.9 & 32.6 & \textbf{50.5} & \textbf{66.0} & \textbf{71.0} & \textbf{30.3} & \textbf{52.4} & \textbf{69.6} & 76.0 & 24.6 \\ \hline
\end{tabular}}
\end{center}
\vspace{-1em}
\label{table:teacher_net}
\end{table*}

\begin{table*}
\caption{Evaluation results of the Student Net on MSMT17, DukeMTMC-reID, and Market1501. We use ResNet50 as our backbone and trained on MSMT17.}
\begin{center}
\resizebox{\textwidth}{!}{
\begin{tabular}{c|c|c|c|c|c|c|c|c|c|c|c|c|c}
\hline
loss & negative set & \multicolumn{4}{c|}{Test on MSMT17} & \multicolumn{4}{c|}{Transfer to DukeMTMC-reID} & \multicolumn{4}{c}{Transfer to Market1501} \\ \hline
\hline
HHL \cite{Zhong_2018_ECCV} & NA & - & - & - & - & 45.0 & 59.4 & 64.4 & 23.0 & \textbf{56.0} & \textbf{75.8} & \textbf{81.2} & \textbf{26.7}\\
\hline
CE-FAT&batchNeg&\textbf{69.4}&\textbf{81.5}&\textbf{85.6}&\textbf{39.2}&49.2&64.8&69.6&28.7&50.6&68.0&74.9&23.6\\
\hline
CE-FAT-distillation&batchNeg & 66.2 & 79.2 & 83.6 & 36.5 & \textbf{50.9} & \textbf{66.6} & \textbf{72.2} & \textbf{31.3}& 52.8 & 69.2 & 75.9 & 25.4 \\
\hline
\end{tabular}}
\end{center}
\vspace{-1.5em}
\label{table:student_net}
\end{table*}

\vspace{-1em}
\paragraph{Implementation of FAT Loss}
We implement our FAT loss in PyTorch deep learning framework. In the training phase, all images are resized to 144$\times$432 and then randomly cropped into 128$\times$384 sub-images. Standard horizontal flipping is adopted for data augmentation. In the test phase, all images are re-sized to 128$\times$384 and no data augmentations are applied. All images have the training set mean subtracted and then normalized by the training set standard deviation, before feeding into the network.

Following a standard ReID protocol, we use ResNet \cite{he2016deep} or Densenet \cite{huang2017densely} backbone as the feature extractor $f_E$ towards learning a pedestrian representation directly supervised by FAT loss $L_\text{fat}$. The cluster centroids are computed at the beginning of each epoch, using $C_{i1}$ for FAT loss and $C_{i4}$ for normalized FAT loss in Equation \ref{center}. Besides, we also compare four different options of choosing the negative cluster $c_n$ for computing FAT loss each time: i) \textit{ctrdAll:} identity classes that are different from the one $a$ belong to; ii) \textit{ctrdAvg:} consider all other classes, except the one that $a$ belongs to, as one cluster and obtain one negative centroid by computing the average of all negative centroids, which is similar to \cite{Yu_2018_ECCV_2} but differs in the way of calculating all negative samples' mean; iii) \textit{ctrdHM:} find a hard negative cluster (in terms of closest centroid to the one that $a$ belongs to), from all classes of the whole dataset; iv) \textit{batchHM:} find a hard negative sample on ``batch level'', \textit{e.g.}, from all classes that are sampled by the current batch.  

\vspace{-1em}
\paragraph{Implementation of Label Distillation} The heavy label noise on MSMT17 further motivates us to conduct label distillation experiments on it. Following the basic routine described in Section 3.2, we further study four different modes of identifying confident samples: i) \textit{hard threshold}: select all samples whose softmax entropy values are below a pre-set threshold $t$ as the \underline{trusted training subset}, and discard all un-selected samples; ii) \textit{soft threshold}: select all samples whose softmax entropy values are below a pre-set threshold $t/2$, and then randomly select 50\% of the remaining (unselected) samples to add into the trusted training subset; iii) \textit{hard percentage}: always select 50\% samples with lowest softmax entropy values, as the trusted training subset; iv) \textit{hard percentage}: always select 25\% samples with lowest softmax entropy values first, and then randomly select another $1/3$ from the remaining 75\% (unselected) samples to add into the trusted training subset. 

\begin{figure}[t]
\begin{center}
\vspace{-1em}
    \includegraphics[width=\linewidth]{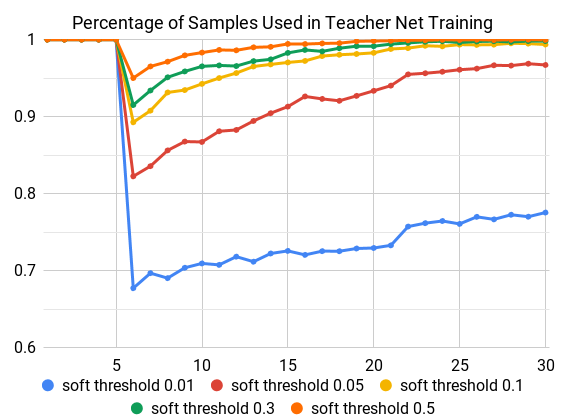}
\end{center}
\vspace{-1.5em}
\caption{The number of samples actually used as the trusted training subset, when training the ResNet50 teacher model with different soft threshold $t$ values, on the Market1501 dataset.}
\vspace{-1em}
\label{fig:tNetTraining}
\end{figure}

The important difference between ``threshold'' and ``percentage'' methods lies in whether we keep a constant or dynamic size of the trusted training subset for the teacher model. For the first two threshold-based methods, even sticking to the same $t$ throughout one training, the portion of samples selected into the trusted set will be dynamic, as more samples might become better confident as training continues. Figure \ref{fig:tNetTraining} visualizes this trend: given $t \le 0.1$, the final training stage will always have considered all training samples as trusted; while a larger $t$ may lead to more ``conservative'' selection. We choose  $t$ = 0.1 as the empirical default value found in experiments for i) and ii). Also, for the two ``soft'' strategies ii) and iv), our hope is to utilize a larger set of samples while letting the stochastic selection ``smooth out'' the impacts from noisy labels.

\subsection{Comparison Analysis on FAT loss}

We first present a comprehensive ablation study on the effectiveness of FAT loss in Table \ref{table:hyperparameters_market}, using the Market1501 dataset. By default, we use the CE-FAT loss defined in (\ref{fat-h}), with  $\lambda$ = 1, as it consistently improves over either FAT or CE loss alone. The margin $m$ is chosen as 1 for FAT loss and 0.1 for normalized FAT loss, as validated to be effective in experiments. We study on the four choices of the negative cluster (only ctrdAvg was previously explored in a similar form \cite{Yu_2018_ECCV_2}), as well as the FAT loss hyperparameter (margin $m$). We also compare CE-FAT with CE-P2S, the latter defined by removing the cluster compactness term in FAT loss; as well as the normalized versions for both, denoted as CE-FATnorm and CE-P2S norm, respectively. 

We evaluate different methods in terms of their top-1/top-5/top-10 accuracy and mean average precision (mAP) values obtained on the Market1501 testing set. Moreover, we use the \textbf{direct transfer} performance of the Market1501-trained feature extraction to the DukeMTMC-reID dataset, as an additional performance criterion, to avoid overfitting small ReID datasets. A few popular ReID loss options proposed in previous works \cite{ustinova2016learning,li2017person,zhou2017point,Alexander_2017} are also included into comparison, so is a CycleGAN \cite{Deng_2018_CVPR} baseline for transfer evaluation. Note that  CycleGAN is a domain adaption method that demands re-training on the target domain, while the direct transfer needs no extra re-training. 

\underline{First}, comparing CE-FAT with ctrdAll, ctrdAvg, ctrdHM, and batchNeg, it is clear that batchNeg outperforms the other three.  
\underline{Second}, comparing CE-P2S with CE-FAT in fair settings, we show the necessity of cluster compactness regularization in addition to the P2S loss; for example, without the compactness term, we will see 1.8\% (ctrdAll) and 2.2\% (batchNeg) top-1 accuracy drops on the Market1501 test case, and 7.5\% (ctrdAll) and 9.2\% d(batchNeg) top-1 accuracy drops on the transfer case to DukeMTMC-reID. The performance gaps clearly differentiate FAT loss from previous empirical P2S losses, thanks to our more rigorous upper-bound derivation. \underline{Third}, no performance gain has been observed on Market1501, when using normalized features for FAT/P2S.
\underline{Finally}, CE-FAT outperforms all state-of-the-art losses trained with the same ResNet50, on the Market1501 testing set. Furthermore, after we replace the backbone into DenseNet161, CE-FAT achieves not only further boosted Market1501 testing results, but also impressive direct transfer performance to DukeMTMC-reID, even surpassing Cycle-GAN domain adaption \cite{Deng_2018_CVPR} that is re-trained with the target domain data.

Tables \ref{table:hyperparameters_duke} and \ref{table:hyperparameters_msmt} report similar experiments using DukeMTMC-reID ad MSMT17 datasets, respectively. With most observations aligned with the Market1501 cases, we find the training behavior on MSMT17 to slightly differ from the other two (much) smaller datasets. In particular, while batchNeg remains effective for its own testing set, ctrdAll becomes the best option when it comes to the feature transferability evaluation. That might be attributed to the heavier label noise on MSMT17, that likely benefits from averaging the triplet effects between with current one and all other clusters. Also, we observe CE-FATnorm to outperform CE-FAT, when transferring from MSMT17 to the other two datasets. That implies that normalization may become essential to overcome feature scale variances on large datasets. Finally, training ResNet50 with CE-FAT loss and batchNeg has surpassed the state-of-the-art performance \cite{Wei_2018_CVPR} ever reported on MSMT17.

\subsection{Effect of Label Distillation}
To overcome the noisy label issue on MSMT17, we next investigate label distillation to further unleash the power of FAT loss. Both teacher and student nets adopt the same ResNet50 backbone for simplicity. 

As shown in Table \ref{table:teacher_net}, for the training of the teacher net, the soft threshold/percentage methods appear to outperform their hard counterparts, as they can learn with a wider variety of samples (while hard methods may tend to select too many similar easy samples), meanwhile smoothing out the negative impacts of potential noisy samples due to stochastic sampling/averaging effects. In comparison, soft threshold seems to produce superior results on the same MSMT17 testing set, whereas soft percentage leads to better feature transferability. It implies that soft percentage suffers from less overfitting, because of its curriculum-style learning (as Figure \ref{fig:tNetTraining} shows) that progressively takes into account the entire dataset information. To our surprise, our teacher net trained with only the trusted subsets by soft threshold/percentage yield competitive or even superior performance than the one trained with the whole dataset, in particular on transfer cases. That proves that the teacher net learns effectively and without being misled by  noisy labels.

We then pick the teacher net trained with soft percentage, due to its best transfer performance, to provide soft pseudo labels for training the student net. The training of student net is supervised by the CE-FAT loss with the batchNeg strategy, using the soft pseudo labels in place of original one-hot labels for both CE and FAT terms. The new model in Table \ref{table:student_net}, dubbed CE-FAT-distillation, does not lead to better test results on MSMT17 than our best result (CE-FAT with batchNeg) in Section 4.2. However, it produces state-of-the-art \textbf{direct transfer} performance from MSMT17 to DukeMTMC-reID. Its transfer performance to Market1501 largely surpasses that of CE-FAT without distillation, and shows competitiveness to state-of-the-art HHL domain adaption \cite{Zhong_2018_ECCV}. To re-iterate, direct transfer does not re-train on target domain data as domain adaption has to.

\section{Conclusion}
This work proposes the fast-approximated triplet (\textbf{FAT}) loss, which remarkably improve the efficiency over the standard triplet loss in ReID models. Instead of using point-to-point distances, the FAT loss uses a point-to-set distances with cluster compactness regularization, which is derived rigorously as an upper bound of standard triplet loss, with linear complexity to the training set size. A distillation network is also designed to assign soft labels for samples in place of potentially noisy hard labels. Extensive experiments demonstrate the high effectiveness and promise of the proposed FAT loss along with label distillation.

{\small
\bibliographystyle{ieee}
\bibliography{egbib}
}

\end{document}


\title{Supplementary Material 974 \\ In Defense of the Triplet Loss Again: Learning Robust Person Re-Identification with Fast Approximated Triplet Loss and Label Distillation}

\author{First Author\\
Institution1\\
Institution1 address\\
{\tt\small firstauthor@i1.org}
\and
Second Author\\
Institution2\\
First line of institution2 address\\
{\tt\small secondauthor@i2.org}
}

\maketitle

\section{t-SNE Visulization of Cross Entropy Loss and FAT Loss}

\begin{figure*}[!ht]
\begin{center}
    \includegraphics[width=0.7\linewidth]{figures/CE.png}
    \includegraphics[width=0.7\linewidth]{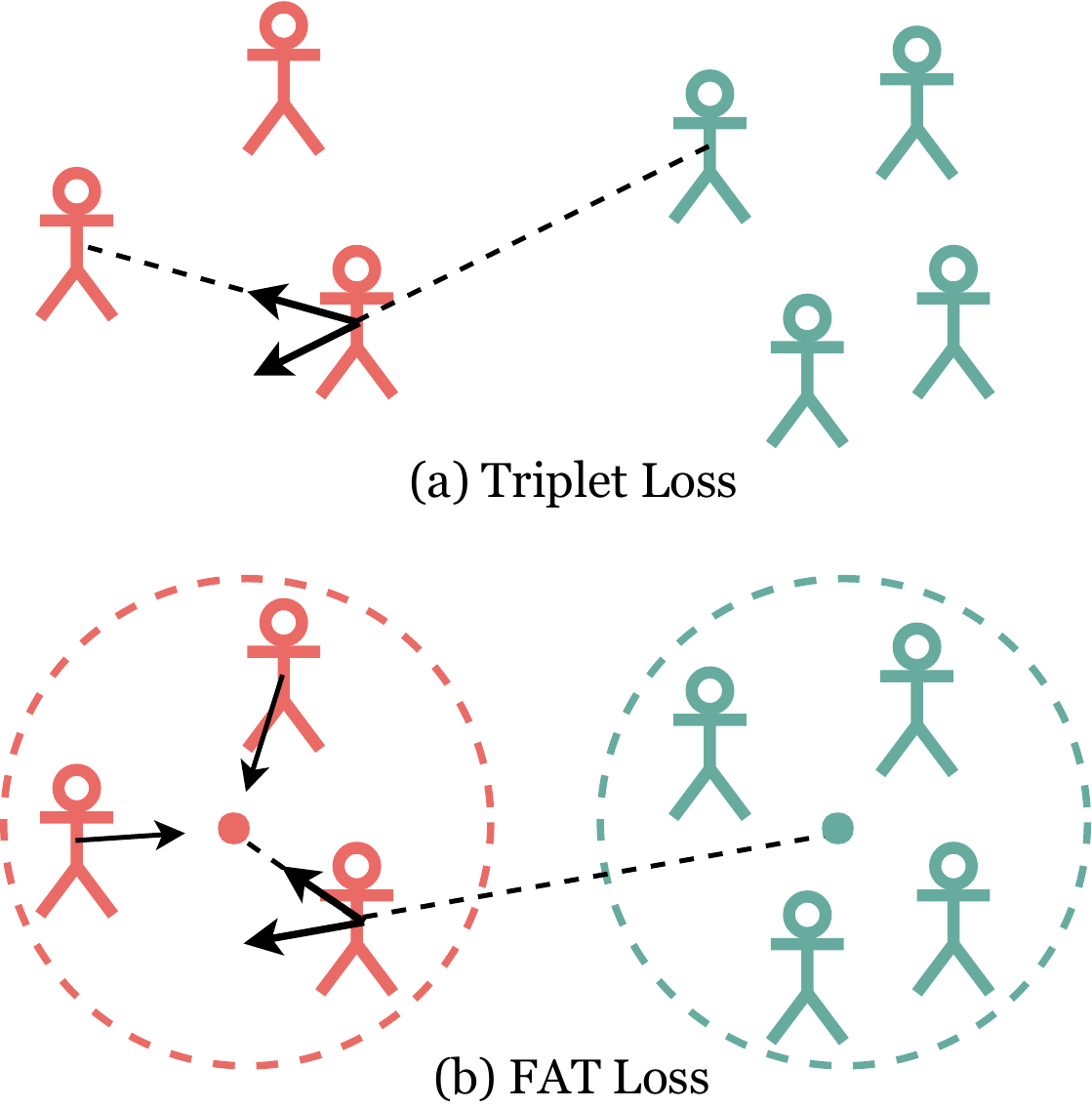}
\end{center}
\vspace{-1em}
\caption{t-SNE visualization of feature distributions, using cross entropy loss (top) and FAT loss (bottom). Twenty identities are randomly selected from the MSMT17 dataset and their IDs are listed below the graphs.}
\vspace{-1em}
\label{fig:tsne}
\end{figure*}

We use t-SNE to visualize the feature distributions learned using cross entropy loss (Figure \ref{fig:tsne} top) and FAT loss (Figure \ref{fig:tsne} bottom). Twenty identities are randomly selected from the MSMT17 dataset and their IDs are listed below the graphs. We can see that the distances between identity features become much larger when we switch from the cross entropy loss to the FAT loss, indicating that the proposed FAT loss is a better optimization target for maximizing the inter-class distance.

\vspace{-1em}

\section{Example of Label Noise}
We empirically found many label noises in the MSMT17 dataset, and Figure \ref{fig:label_noise} shows some fail test cases due to label noises. The leftmost column lists three query images, and the remaining 10 columns shows top-10 retrieval results. Based on the labels provided in the dataset, all the top-10 retrieval images do not belong to the same person in the query image.

We observed three typical cases of label noise: 1) label flip (first row in Figure \ref{fig:label_noise}, the query label should be 1748 but 1749 is provided), where the correct retrievals are determined as false positives due to the wrong query label; 2) mislabel (second row), where the query image contains incomplete person and lacks necessary information for a successful retrieval; 3) coexistence (third row), where the query contains multiple people and makes the model fail to learn an accurate representation. These label noises provided incorrect information and therefore significantly damage the model's learning process.

\begin{figure*}[!ht]
\begin{center}
    \includegraphics[width=\linewidth]{figures/label_noise.pdf}
\end{center}
\vspace{-1em}
\caption{Examples of label noises: label flip (first row), mislabel (second row) and coexistence (third row). The images with red border are all false positive retrieval results due to the label noises.}
\vspace{-1em}
\label{fig:label_noise}
\end{figure*}

\newpage

\section{Label Distillation Pipeline}
Overall, the model is composed of two components, the feature extractor and a classifier. The feature extractor learns a robust representation of the person images while the auxiliary classifier computes the cross entropy loss to supervise training.

We first use a self-bootstrapping approach to learn the teacher model robustly. The teacher net is first trained with all samples (including noisy labels) using the cross entropy loss for 5 epochs. We then resume training for the remaining epochs, but in each epoch, we keep only those identified confident samples and discard samples that are more likely to be outliers or contain label noise. The sample noise is selected by the classifier's prediction, where a cross entropy value larger than a predefined threshold 0.2 is considered to be noisy label sample or outliers. We periodically repeat the above process, and gradually enlarge the pool of confident examples as the training proceeds. 

After the teacher model is trained, its predictions are treated as soft labels to replace the original labels, for training the student model with FAT loss. Only the ``confident'' labels selected by the teacher net will be averaged for cluster centroids estimation. If we use the hybrid FAT loss, then soft labels are the prediction targets for the cross entropy loss too. 

\begin{figure*}[!ht]
\begin{center}
    \includegraphics[width=\linewidth]{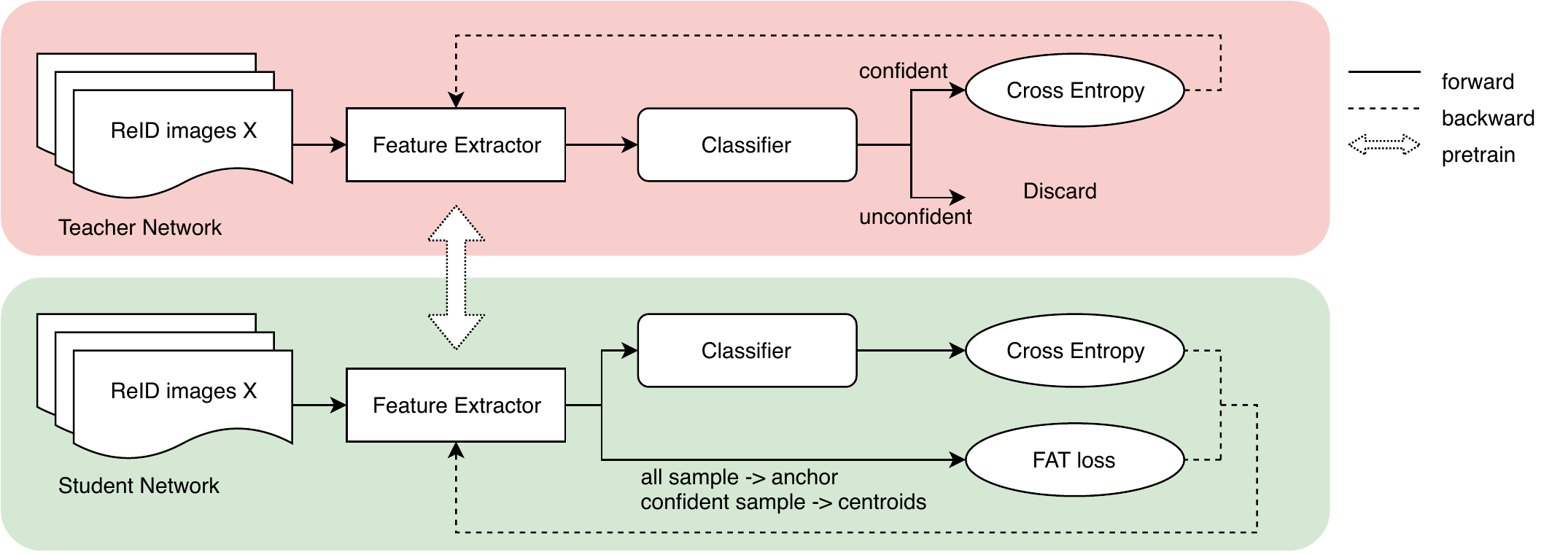}
\end{center}
\vspace{-1em}
\caption{Overview of the proposed label distillation pipeline.}
\vspace{-1em}
\label{fig:model1}
\end{figure*}